\documentclass[journal]{IEEEtran}

\usepackage{graphicx}
\usepackage{amssymb}
\usepackage{amsmath}
\usepackage{multirow}
\usepackage{cite}
\usepackage{bm}
\usepackage{booktabs}
\usepackage{caption}
\usepackage{color}

\author{Zhihao Chen\inst{1} \and
Yang Zhou\inst{2} \and
Anh Tran\inst{3} \and
Junting Zhao\inst{1} \and
Liang Wan\inst{1} \and
Gideon Ooi\inst{3} \and
Lionel Cheng\inst{3} \and
Choon Hua Thng\inst{3} \and
Xinxing Xu\inst{2} \and 
Yong Liu \inst{2} \and
Huazhu Fu\inst{2}}
\begin{document}
\title{Medical Phrase Grounding with Region-Phrase Context Contrastive Alignment} 
\author{Zhihao Chen$^{\star}$, Yang Zhou$^{\star}$, Anh Tran, Junting Zhao, Liang Wan$^{\dag}$, Gideon Ooi, Lionel Cheng, Choon Hua Thng, Xinxing Xu, Yong Liu and Huazhu Fu$^{\dag}$
\thanks{Z.~Chen, J.~Zhao, and L.~Wan are with the College of Intelligence and Computing, Tianjin University, Tianjin 300350, China.}
\thanks{Y.~Zhou, X.~Xu, Y.~Liu, and H.~Fu are with the Institute of High Performance Computing (IHPC), Agency for Science, Technology and Research (A*STAR), 1 Fusionopolis Way, \#16-16 Connexis, Singapore 138632, Republic of Singapore.}
\thanks{A.~Tran is with SGH Department of Diagnostic Radiology.}
\thanks{L.~Cheng is with Saw Swee Hock School of Public Health, National University of Singapore National University Health System, Block MD3, 03–20, 16 Medical Drive Singapore 117597, Singapore.}
\thanks{G.~Ooi and C.~Thng are with the Department of Nuclear Medicine and Molecular Imaging, Division of Radiological Sciences, Singapore General Hospital.}
\thanks{Z.~Chen and Y.~Zhou are the co-first authors.}
\thanks{L.~Wan and H.~Fu are the co-corresponding authors. }
}

\maketitle

\begin{abstract}
Medical phrase grounding (MPG) aims to locate the most relevant region in a medical image, given a phrase query describing certain medical findings, which is an important task for medical image analysis and radiological diagnosis. 
However, existing visual grounding methods rely on general visual features for identifying objects in natural images and are not capable of capturing the subtle and specialized features of medical findings, leading to a sub-optimal performance in MPG.
In this paper, we propose \textbf{MedRPG}, an end-to-end approach for MPG. MedRPG is built on a lightweight vision-language transformer encoder and directly predicts the box coordinates of mentioned medical findings, which can be trained with limited medical data, making it a valuable tool in medical image analysis. 
To enable MedRPG to locate nuanced medical findings with better region-phrase correspondences, we further propose \textbf{T}ri-\textbf{a}ttention \textbf{Co}ntext contrastive alignment (\textbf{TaCo}). 
TaCo seeks \textit{context alignment} to pull both the features and attention outputs of relevant region-phrase pairs close together while pushing those of irrelevant regions far away. This ensures that the final box prediction depends more on its finding-specific regions and phrases. 
Experimental results on three MPG datasets demonstrate that our MedRPG outperforms state-of-the-art visual grounding approaches by a large margin. Additionally, the proposed TaCo strategy is effective in enhancing finding localization ability and reducing spurious region-phrase correlations.
\end{abstract}


\IEEEpeerreviewmaketitle

\section{Introduction}

\begin{figure*}[!t]
\centering
\includegraphics[width=1\textwidth]{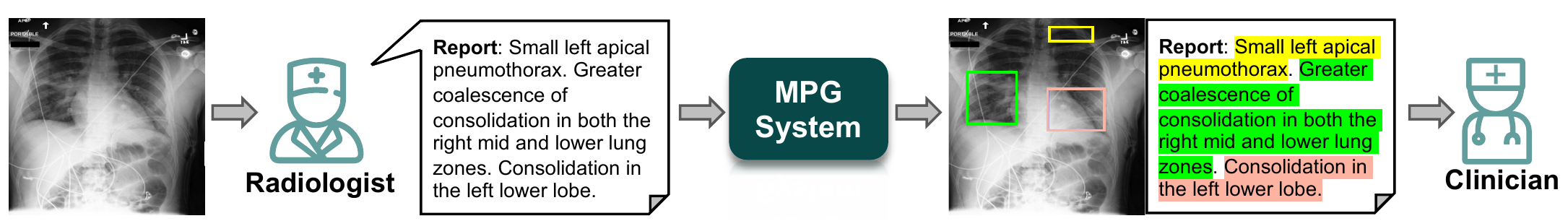}
\caption{Illustration on how MPG helps radiological diagnosis.}
\label{fig:intro}  
\end{figure*}

Medical phrase grounding (MPG) is the task of associating text descriptions with corresponding regions of interest (ROIs) in medical images. It enables machines to understand and interpret medical findings mentioned in medical reports in the context of medical images, which is crucial in medical image analysis and radiological diagnosis. Figure~\ref{fig:intro} illustrates how an MPG system facilitates the radiological diagnosis process. Radiologists first review the medical images (\textit{e.g.}, X-rays, CT scans, and MRI scans) to find out possible abnormalities and then write a report that summarizes their findings. Then, given the image and report, the MPG system can help doctors to locate and link ROIs to the corresponding phrases in the reports, which reduces the time of the diagnostic process and improves the quality of risk stratification and treatment planning. 

In this paper, we study the MPG problem and focus on a typical setting to learn the grounding between Chest X-ray images and medical reports. As far as we know, there are only a few related works on the medical phrase grounding problem. (This is probably because medical annotations of grounding data require specialized expertise and are time-consuming and expensive to be collected.) Benedikt \textit{et al.}~\cite{boecking2022making} made use of text semantics to improve biomedical vision-language processing. They first evaluated the grounding performance of self-supervised biomedical vision-language models by proposing an MPG benchmark. However, their focus is on vision-language pre-training rather than addressing the MPG problem. Qin \textit{et al.}~\cite{qin2022medical} proposed to transfer the knowledge of general vision-language models for detection tasks in medical domains. The key idea is to guide the vision-language model through hand-crafted prompting of visual attributes such as color, shape, and location that may be shared between natural and medical domains. This approach fails to consider unique characteristics in radiological images and reports and is inapplicable to MPG for radiological images.

Although visual grounding has been well studied for natural images~\cite{deng2021transvg,yu2018mattnet,chen2022multi,zhu2022seqtr,yang2019fast}, it is non-trivial to apply these approaches to radiological images. Specifically, MPG requires learning specialized visual-textual features so that the model can identify medical findings with subtle differences in texture and shape and interpret the relative positions mentioned in the medical reports. In contrast, general grounding methods often rely on visual features that are useful for object detection or classification but not specific to medical images, leading to inaccurate region-phrase correlations and thus sub-optimal results. In addition, many grounding models for general domains are too heavy to be trained with limited annotated data, which is common. Such heavy model structures are generally difficult to be trained with limited annotated data.

In this work, we propose \textbf{MedRPG}, an end-to-end approach for MPG. MedRPG has a lightweight model architecture and explicitly captures the finding-specific correlations between ROIs and report phrases. Specifically, we propose to stack a few vision-language transformer layers to encode both the medical images and report phrases and directly predict the box coordinates of desired medical findings. Compared to general grounding methods with heavy model architectures, this design is more robust against overfitting for MPG with limited training data. To locate nuanced medical findings with better region-phrase correspondences, we further propose \textbf{T}ri-\textbf{a}ttention \textbf{Co}ntext contrastive alignment (TaCo). TaCo seeks \textit{context alignment} to learn finding-specific representations that jointly align the region, phrase, and box prediction under the same context of a vision-language transformer encoder. It pulls both the \textit{features} and \textit{attention outputs} close together for semantically relevant region-phrase pairs while pushing those of irrelevant pairs far away. This encourages the alignment between regions and phrases at both feature and attention levels, leading to enhanced finding-identification ability and reduced spurious region-phrase correlations. Experimental results on three medical datasets demonstrate that our MedRPG is more effective in localizing medical findings, achieves better region-phrase correspondences, and significantly outperforms general visual grounding approaches on the MPG task. \textbf{We will release all codes after acceptance.}



\section{Proposed Method}
\begin{figure*}[!t]
\centering
\includegraphics[width=1\textwidth]{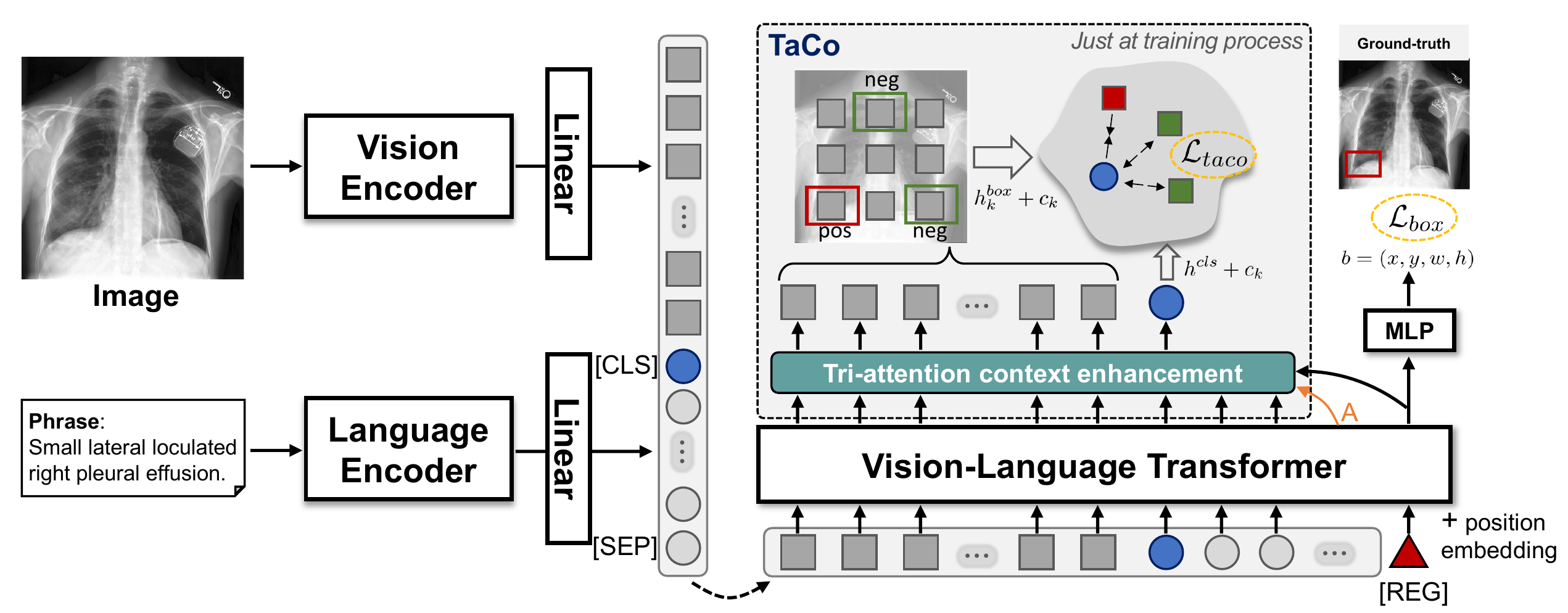}
\caption{An overview of the proposed MedRPG method.}
\label{fig:framework}
\end{figure*}
The MPG problem can be defined as follows: Given a radiological image $\mathbf{I}$ associated with medical phrases $\mathbf{T}$ written by specialist radiologists, MPG aims to locate the described findings and then output a 4-dim bounding box (bbox) coordinates $\bm{b}=(x,y,w,h)$, where $(x,y)$ is the box center coordinates and $w,h$ are the box width and height, respectively.

\subsection{Model Architecture} \label{sec:overview}

Fig~\ref{fig:framework} illustrates the framework of our method. Given image $\mathbf{I}$ and phrase $\mathbf{T}$, we first leverage the Vision Encoder and Language Encoder to generate the image and text embeddings. Next, we concatenate the multi-modal feature embeddings and append a learnable token (named [REG] token), and then feed them into a lightweight Vision-Language Transformer to encode the intra and inter-modality context in a common semantic space. Finally, the output state of the [REG] token is employed to predict the 4-dim bbox via a grounding head. Additionally, to ensure a consistent representation of medical findings across modalities, we introduce TaCo, which aligns the context of region and phrase embeddings at both the feature and attention levels.



\noindent\textbf{Vision Encoder}: Following the common practice \cite{deng2021transvg,kamath2021mdetr,carion2020end}, the visual encoder starts with a CNN backbone, followed by the visual transformer. We choose the ResNet-50~\cite{he2016deep} as the CNN backbone. The visual transformer includes 6 stacked transformer encoder layers~\cite{dosovitskiy2020vit}. Given a radiological image $\mathbf{I} \in \mathbb{R}^{3 \times W \times H}$, it is fed into the CNN backbone to obtain the high-level deep features. Next, we apply a $1 \times 1$ ConV layer to project the deep features into a $C_v$-dimensional subspace. Finally, we exploit the visual transformer to mine the long-range visual relations and further output the visual features $\mathbf{F}_{v}=[\bm{f}_v^n]_{n=1}^{N_v}$, where $N_v$ is the number of visual tokens and $\bm{f}_v^n \in \mathbb{R}^{C_v}$ is the $n$-th token of $\mathbf{F}_{v}$.


\noindent\textbf{Language Encoder}: We leverage pre-trained language models such as BERT~\cite{kenton2019bert} as the language encoder, which includes 12 transformer encoder layers. Given a medical phrase $\mathbf{T}$, we first utilize the BERT tokenizer to convert it into a sequence of tokens. Next, we follow the common practice to append a [CLS] token at the beginning as the global representation of the input medical phrases and append a [SEP] token at the end, and then pad the sequence to a fixed length. Finally, we use BERT to encode the tokens into the text embeddings $\mathbf{F}_{l}=[\bm{f}^{cls}, \bm{f}_l^n]_{n=2}^{N_l}$, where $N_l$ is the number of text tokens, $C_l$ is the feature dimensions, and $\bm{f}_l^n \in \mathbb{R}^{C_l}$ is the $n$-th token of $\mathbf{F}_{l}$.


\noindent\textbf{Vision-Language Transformer}: After the individual vision and language encoding, we obtain $\mathbf{F}_{v}$ and $\mathbf{F}_{l}$. To capture the correspondence between the image and phrase embeddings, we first project them into the common space (channel= $C_{vl}$) and then fed them into a Vision-Language Transformer (VLT), together with an extra learnable [REG] token, which is further used to predict the bbox:
\begin{equation}\label{Equ:VLT}
    \mathbf{H} = \mathrm{VLT} \left( [\varphi_v(\mathbf{F}_{v} ), \varphi_l(\mathbf{F}_{l}), \bm{r}]\right),
\end{equation}
where $\varphi_v(\cdot)$ and $\varphi_l(\cdot)$ denote the project functions for vision and language tokens, respectively. $\bm{r}$$\in$$\mathbb{R}^{C_{vl}}$ is the [REG] token and $\mathrm{VLT}(\cdot)$ denotes the VLT encoder with learnable position embeddings. $\mathbf{H} \in \mathbb{R}^{C_{vl} \times N_{vl}}$ (where $N_{vl}=N_v+N_l+1$) is the output of VLT that consist of three parts: vision embeddings $\mathrm{H}_v$$=$$[\bm{h}_v^n]_{n=1}^{N_v}$, language embeddings $\mathrm{H}_l$$=$$[\bm{h}^{cls}, \bm{h}_l^n]_{n=2}^{N_l}$, and [REG] embedding $\bm{h}^{reg}$. 



To perform the final grounding results, we further feed $\bm{h}^{reg}$ into a 3-layer MLP to predict the final 4-dim box coordinates $\bm{b}=\mathrm{MLP}(\bm{h}^{reg})$. Given the grounding-truth box $\bm{b}_0$, we leverage smooth L1 loss~\cite{girshick2015fast} and GIoU~\cite{rezatofighi2019generalized} loss which are popular in grounding and detection tasks to optimize our model:
\begin{equation}\label{Equ:box_loss}
    \mathcal{L}_{box} = \mathrm{\Phi}_{l1}(\bm{b}, \bm{b}_0) + \lambda \cdot \mathrm{\Phi}_{giou}(\bm{b}, \bm{b}_0),
\end{equation}
where $\mathcal{L}_{box}$ is the box loss. $\mathrm{\Phi}_{l1}$ and $\mathrm{\Phi}_{giou}$ are the smooth L1 and GIoU loss functions, respectively. $\lambda$ is the trade-off parameter.

\subsection{Tri-attention Context Contrastive Alignment}\label{sec:eca}
Medical findings often share subtle differences in texture and brightness level due to the low contrast of the medical images, which makes it challenging for the MPG methods to capture accurate region-phrase correlations. To identify nuanced medical findings with better region-phrase correspondences, we propose the Tri-attention Context Contrastive Alignment (TaCo) strategy to learn finding-specific representations with accurate region-phrase correlations by explicitly aligning relevant regions and phrases at both feature and attention levels.

\noindent\textbf{Feature-Level Alignment.} The feature-level alignment aims to make visual and textual embeddings with the same semantics meaning to be similar. To this end, given the bbox $\bm{b}_0$ related to a given phrase query, we first obtain the positive ROI embeddings $\bm{h}^{box}_{0} \in \mathbb{R}^{C_{vl}} =\textrm{Pool}(\mathbf{H}_v, \bm{b}_0)$ by aggregating visual embeddings $H_v$ within the bbox $\bm{b}_0$. Next, we randomly select $K$ bbox $\{\bm{b}_k\}_{k=1}^{K}$ that have low IoUs with $\bm{b}_0$ (i.e., regions that are irrelevant to the given phrase query) and obtain $K$ negative region embeddings $\{\bm{h}^{box}_{k} \in \mathbb{R}^{C_{vl}}\}_{k=1}^{K}$. Let $\bm{h}^{cls}$ be the features of the input phrases. We want to make the positive ROI embedding $\bm{h}^{box}_{0}$ close to the corresponding phrase embedding $\bm{h}^{cls}$ whereas negative region embeddings $\{\bm{h}^{box}_{k}\}_{k=1}^{K}$ far away. This is achieved by exploiting the InfoNCE~\cite{oord2018representation,gutmann2010noise} loss as:
\begin{equation}\label{Equ:vanilla_ca_loss}
    \mathcal{L}_{fea} = -\textrm{log}\frac{\textrm{exp}(\bm{h}^{cls} \cdot \bm{h}^{box}_{0} / \tau)}{\sum_{k=0}^{K}{\textrm{exp}(\bm{h}^{cls} \cdot \bm{h}^{box}_{k} / \tau)}},
\end{equation}
where $\mathcal{L}_{fea}$ denotes the feature-level alignment loss, $\tau$ is a temperature hyper-parameter and `$\cdot$' represents the inner (dot) product.

\noindent\textbf{Attention-Level Alignment.}
In addition to the feature-level alignment, we also consider attention-level alignment, which encourages the attention outputs of VLT for relevant region-phrase pairs to be similar. To realize this, we extract the attention weight $\mathbf{A} \in \mathbb{R}^{N_{vl} \times N_{vl}}$ from the last multi-head attention layer of VLT. We denote $\bm{a}^{reg}$, $\bm{a}^{cls}$ and $\{\bm{a}^{box}_{k}\}_{k=0}^K$ as the attention weights for the embeddings of the [REG] token, the [CLS] token, and the $K+1$ bboxes, respectively, where $\bm{a}^{box}_{k} = \mathtt{Pool}(\mathbf{A}, \bm{b}_k)$. Given the $k$-th bbox embedding, we calculate the joint attention weights of bbox, [CLS], and [REG] embeddings and then further product $\mathbf{H}$ to get the triple-attention context pooling $\bm{c}_k$ as follows:
\begin{equation}\label{Equ:cal_t}
    \bm{c}_k = \sum_{j=0}^{N_{vl}}{\left( \bm{t}_k^{(j)} \cdot \mathbf{H}[:,j] \right)}, \;\; \text{where}~\bm{t}_k = \mathtt{Norm}(\bm{a}^{cls} \cdot \bm{a}^{reg} \cdot \bm{a}^{box}_{k}),
\end{equation}
where $\bm{t}_k$ represents the joint attention weights, $\bm{t}_k^{(j)}$ denotes the $j$-th element of $\bm{t}_k$, $\mathbf{H}[:,j]$ denotes the $j$-th column of $\mathbf{H}$, and $\mathtt{Norm}(\cdot)$ is the L2 normalization operation to constrain the sum of the squared weights to be equal to 1. Such triple-attention context pooling $\bm{c}_k$ characterizes the contextual dependencies among regions, phrases, and box predictions in the VLT. Intuitively, the box prediction of certain medical findings should be made based on its relevant regions and phrases rather than irrelevant ones. Therefore, the attention outputs $\bm{c}_0$ for relevant region-phrase pairs should be similar to their individual embeddings $\bm{h}^{box}_{0}$ and $\bm{h}^{cls}$, leading to attention-level alignment.


\begin{table*}[!t]
\footnotesize
\centering
\setlength{\tabcolsep}{3.6pt}
\caption{Grounding results on MS-CXR~\cite{boecking2022making}, ChestX-ray8~\cite{wang2017chestx}, and the in-house datasets with respect to Acc and mIoU.}
  \resizebox{0.75\textwidth}{!}{
\begin{tabular}{c|c|c|cc|cc|cc}
\toprule
 \multirow{2}{*}{Method} & Vision & Language & \multicolumn{2}{c|}{MS-CXR} & \multicolumn{2}{c|}{ChestX-ray8} & \multicolumn{2}{c}{In-house} \\
  & Encoder & Encoder & Acc$\uparrow$  & mIoU$\uparrow$  & Acc$\uparrow$      & mIoU$\uparrow$ & Acc$\uparrow$      & mIoU$\uparrow$   \\ \toprule
RefTR~\cite{li2021referring} & \multirow{4}{*}{ResNet-50} & BERT & 53.69 & 50.11 & 29.27 & 29.59 & 46.03   & 40.99 \\
VGTR~\cite{du2022visual}  & & Bi-LSTM & 60.27 & 53.58 & 32.65 & \underline{34.02} & 47.37   & 41.92 \\
SeqTR~\cite{zhu2022seqtr} & & Bi-GRU & 63.20 & 56.63 & 32.88 & 33.09 & 44.42 & 39.45  \\
TransVG~\cite{deng2021transvg} & & BERT & \underline{65.87} & \underline{58.91} & \underline{34.51} & 33.98 & \underline{48.30} & \underline{43.35} \\
\midrule
Ours  & ResNet-50 &BERT & \textbf{69.86}  & \textbf{59.37} & \textbf{36.02} & \textbf{34.59}  & \textbf{49.87} & \textbf{43.86}  \\ 
\bottomrule
\end{tabular}
}
\label{table:sota}
\end{table*}

\noindent\textbf{Context Contrastive Alignment.} With the above results, we can propose our TaCo strategy by integrating both feature and attention-level alignments. Specifically, we modify the InfoNCE loss (\ref{Equ:vanilla_ca_loss}) to simultaneously perform feature and attention-level alignments by adding the triple-attention context pooling $\bm{c}_k$ to the respective region and phrase features (i.e., $\bm{h}^{cls}$, $\bm{h}^{box}_k$). This leads to the TaCo loss as follows:
\begin{equation}\label{Equ:ca_loss_lcp}
    \mathcal{L}_{taco} = -\textrm{log}\frac{\textrm{exp}((\bm{h}^{cls}+\bm{c}_0) \cdot (\bm{h}^{box}_{0}+\bm{c}_0) / \tau)}{\sum_{k=0}^{K}{\textrm{exp}((\bm{h}^{cls}+\bm{c}_k) \cdot (\bm{h}^{box}_{k}+\bm{c}_k) / \tau)}}.
\end{equation}

Finally, we combine the TaCo loss (\ref{Equ:ca_loss_lcp}) and box loss (\ref{Equ:box_loss}) to get the overall loss functions of MedRPG:
\begin{equation}\label{Equ:total_loss}
    \mathcal{L}_{MedRPG} = \mathcal{L}_{box} + \mu \cdot \mathcal{L}_{taco}.
\end{equation}
where $\mathcal{L}_{MedRPG}$ denotes total loss for MedRPG and $\mu$ is the trade-off parameter.

\section{Experiments}
\noindent\textbf{Dataset.} Our experiments are conducted on two public datasets, i.e., MS-CXR~\cite{boecking2022making}, ChestX-ray8~\cite{wang2017chestx}, and one in-house dataset~\footnote{The ethical approval of this dataset was obtained from the Ethical Committee.}. MS-CXR is sourced from MIMIC-CXR~\cite{johnson2019mimic,mimic_PhysioNet} and consists of 1,153 samples of Image-Phrase-BBox triples. We pre-process MS-CXR to make sure a given phrase query corresponds to only one bounding box, which results in 890 samples from 867 patients. ChestX-ray8 is a large-scale dataset for diagnosing 8 common chest diseases, of which 984 images with pathology are provided with hand-labeled bounding boxes. Due to the lack of finding-specific phrases from medical reports, we use category labels as the phrase queries to build the Image-Report-BBox triples for our task. Our in-house dataset comprises 1,824 Image-Phrase-BBox samples from 635 patients, including 23 categories of chest abnormalities with more complex phrases. For a fair comparison, both datasets are split into train-validation-test sets by 7:1:2 based on the patients.


\noindent\textbf{Evaluation Metrics.} To evaluate the quality of the MPG task, we follow the standard protocol of nature image grounding~\cite{deng2021transvg} to report Acc(\%), where a predicted region will be regarded as a positive sample if its intersection over union (IoU) with the ground-truth bounding box is greater than 0.5. Besides, we also report mIoU (\%) metric for a more comprehensive comparison.


\begin{table}[!t]
\footnotesize
\centering
\caption{Ablation study on MS-CXR dataset.}
\setlength{\tabcolsep}{3pt}
\begin{tabular}{c|ccccc}
    \toprule
        Metrics & Only images & Only phrases & Baseline & w/ Feature-level & w/ TaCo
    \\
\midrule
Acc $\uparrow$ & 41.91 & 40.12 & 66.86 & 67.26 & \textbf{69.86} \\
mIoU $\uparrow$ & 38.44 & 39.34 & 58.93 & 59.25 & \textbf{59.37} \\
\bottomrule
\end{tabular}
\label{table:ablation}
\end{table}

\noindent\textbf{Baselines.} We compare our MedRPG with SOTA methods for general visual grounding, such as RefTR~\cite{li2021referring}, TransVG~\cite{deng2021transvg}, VGTR~\cite{du2022visual}, and SeqTR~\cite{zhu2022seqtr}. We choose their official implementations for a fair comparison. Since the medical datasets are too small to train a data-hungry transformer-based model from scratch, we initialize our MedRPG (encoders) from the general grounding models pre-trained on natural images. The compared methods share the same settings.

\noindent\textbf{Implementation Details.} The experiments are conducted on the PyTorch~\cite{NEURIPS2019_9015} platform with an NVIDIA RTX 3090 GPU.
The input image size is 640$\times$640. The channel numbers $C_{v}$, $C_{l}$, and $C_{vl}$ are 256, 768, and 256. The sample number $K$ is set to 5. The trade-off parameter $\lambda$ in Eqn.~\ref{Equ:box_loss} and $\mu$ in Eqn.~\ref{Equ:total_loss} are set to 1 and 0.05, respectively. The base learning rates for the vision encoder, language encoder, and vision-language transformer are set to $1$$\times$$10^{-5}$, $1$$\times $$10^{-5}$, and $5$$\times$$10^{-5}$, respectively. We train our MedRPG model by the AdamW~\cite{loshchilov2017decoupled} optimizer for 90 epochs with a learning rate dropped by a factor of 10 after 60 epochs. For all the baselines and MedRPG, we select the best checkpoint for testing based on validation performance and report the average performance metrics computed by repeating each experiment with three different random seeds.
\begin{figure*}[!t]
\centering
\includegraphics[width=1\textwidth]{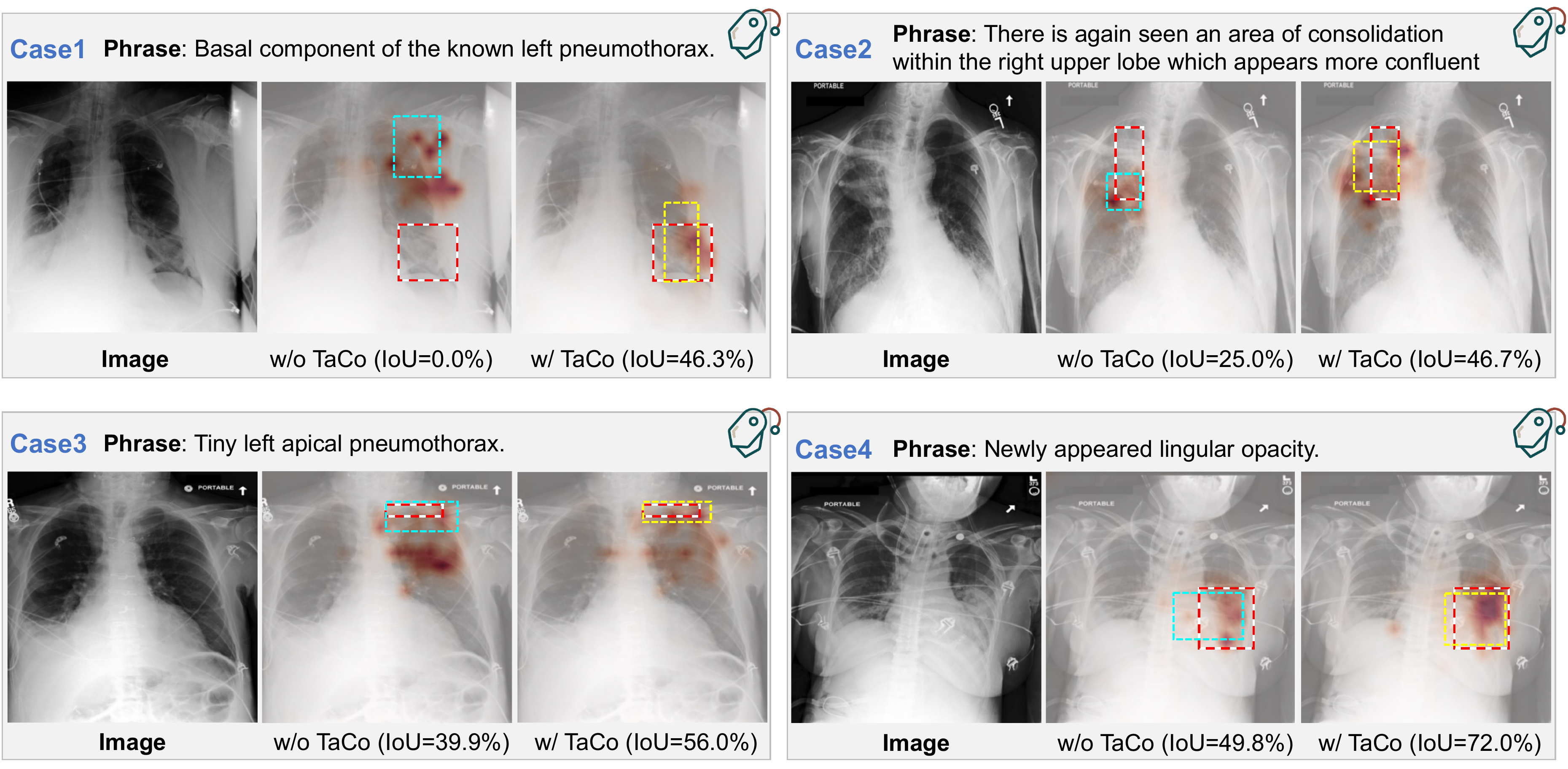}
\caption{Visualized grounding results for MedRPG w/ and w/o TaCo. We show the ground-truth box (\textcolor{red}{red} box), prediction box (\textcolor{cyan}{cyan} or \textcolor{yellow}{yellow} box), and the [REG] token's attention to visual tokens (a heatmap with high values in red).} 
\label{fig:visual}
\end{figure*}

\noindent\textbf{Experimental Results.} Table~\ref{table:sota} provides the grounding results on the MS-CXR, ChestX-ray8, and in-house datasets. As can be seen, our MedRPG consistently achieves the best performance in all cases. In particular, we note that lightweight models like TransVG and our MedRPG generally perform better, which indicates lightweight models are more applicable for MPG. Despite this, our method still outperforms TransVG by a margin of 6.1\% in Acc on MS-CXR. This can be attributed to the proposed TaCo strategy in learning finding-specific representations and improving region-phrase alignment. On ChestX-ray8, all methods get degraded results due to the lack of position cues in the phrase queries. Nevertheless, our method still outperforms the second-best method by 4.3\% in Acc and 1.6\% in mIoU. On the in-house dataset, our method is still the best even when there exist much more types of findings to be grounded.

\noindent\textbf{Ablation Study.}
We conduct ablative experiments on the MS-CXR dataset to verify the effectiveness of each component in MedRPG. Table~\ref{table:ablation} shows the quantitative results of each combination. To verify how the vision and language modalities contribute to the MPG performance, we perform MedRPG with either image or test inputs. As expected, MedRPG can only achieve poor results under the unimodal setting.
Next, we consider the inputs with both images and phrases and observe a significant improvement in performance compared to MedRPG trained from a single modality. Then, we equip MedRPG with feature-level alignment and gain the improvement of 0.6\% in Acc and 0.5\% in mIoU, which suggests it is helpful but still not good enough to learn the accurate region-phrase correspondences. Finally, with the proposed TaCo, MedRPG further gains a significant improvement by 3.8\% in Acc. This shows that TaCo is effective in improving the MPG performance with better region-phrase correlations.

\noindent\textbf{Qualitative Results.} In Figure~\ref{fig:visual}, we show the box predictions and attention maps obtained by MedRPG with and without TaCo to demonstrate the effectiveness of TaCo in identifying abnormal medical findings and capturing region-phrase correlations. For instance, in Case 1, pneumothorax is present in an uncommon location (i.e., lower left lung) and the phrase does not provide an accurate location cue. Without TaCo, MedRPG overfits the upper lung regions where pneumothorax appears more frequently. In contrast, with TaCo, the model can better learn pneumothorax representations and identify the corresponding ROI even without accurate location information. In other cases, although the method without TaCo can also roughly find the location of the medical findings, MedRPG with TaCo can obtain more focused attention maps on the medical findings. It suggests that TaCo is effective in reducing spurious region-phrase correlations, leading to more accurate and interpretable bbox predictions.

\section{Conclusion}
 This study introduces MedRPG, a lightweight and efficient method for medical phrase grounding. A novel tri-attention context contrastive alignment (TaCo) is proposed to learn finding-specific representations and improve region-phrase alignment in feature and attention levels. Experimental results show that MedRPG outperforms existing visual grounding methods and achieves more consistent correlations between phrases and mentioned regions. 
\bibliographystyle{IEEEtran}
\bibliography{mybibliography}

\end{document}